\documentclass{article}

\usepackage{arxiv}

\usepackage[utf8]{inputenc} % allow utf-8 input
\usepackage[T1]{fontenc}    % use 8-bit T1 fonts
\usepackage{hyperref}       % hyperlinks
\usepackage{url}            % simple URL typesetting
\usepackage{booktabs}       % professional-quality tables
\usepackage{amsfonts}       % blackboard math symbols
\usepackage{nicefrac}       % compact symbols for 1/2, etc.
\usepackage{microtype}      % microtypography
\usepackage{lipsum}
\usepackage{color}
\usepackage{times}
\usepackage{epsfig}
\usepackage{graphicx}
\usepackage{amsmath}
\usepackage{amssymb}
\usepackage{subfig}
\usepackage{array}
\usepackage{textcomp}
\usepackage{pdfpages}
\title{UDFNet: Unsupervised Disparity Fusion with Adversarial Networks}

\author{
  Can Pu \\
   \And
Robert B. Fisher \\
}

\begin{document}
\maketitle

\begin{abstract}
Existing disparity fusion methods based on deep learning achieve state-of-the-art performance, but they require ground truth disparity data to train. As far as I know, this is the first time an unsupervised disparity fusion not using ground truth disparity data has been proposed.   

In this paper, a mathematical model for disparity fusion is proposed to guide an adversarial network to train effectively without ground truth disparity data. The initial disparity maps are inputted from the left view along with auxiliary information (gradient, left \& right intensity image) into the refiner and the refiner is trained to output the refined disparity map registered on the left view. The refined left disparity map and left intensity image are used to reconstruct a fake right intensity image. Finally, the fake and real right intensity images (from the right stereo vision camera) are fed into the discriminator. In the model, the refiner is trained to output a refined disparity value close to the weighted sum of the disparity inputs for global initialisation. Then, three refinement principles are adopted to refine the results further. (1) The reconstructed intensity error between the fake and real right intensity image is minimised. (2) The similarities between the fake and real right image in different receptive fields are maximised. (3) The refined disparity map is smoothed based on the corresponding intensity image.   

The adversarial networks' architectures are effective for the fusion task. The fusion time using the proposed network is small. The network can achieve 90 fps using Nvidia Geforce GTX 1080Ti on the Kitti2015 dataset when the input resolution is 1242 * 375 (Width * Height) without downsampling and cropping. The accuracy of this work is equal to (or better than) the state-of-the-art supervised methods. 
\end{abstract}

% keywords can be removed

\section{Introduction}
With the popularity of the disciplines related to 3D vision (eg: robotics, augmented and mixed reality, autonomous driving), how to get more accurate depth information using cheap devices in a 3D environment is important. Currently, there are many methods to obtain depth information, such as active illumination devices (eg: structured light cameras, Time of Flight (ToF) sensors), passive methods (monocular vision~\cite{Monodepth}, stereo vision~\cite{PSMNet,Dispnet,FPGA_stereo,SGM}) and so on. However, none of these methods is perfect in all scenes. For example: ToF is sensitive to sunshine outdoors and reflectivity of the materials; Vision-based methods are sensitive to scene content (repetitive or textureless regions); Lidar-based devices are expensive and data is sparse and lacks color information. Thus, depth fusion from multiple sources is urgently needed, where different data sources can compensate for the weaknesses of each other. 

In recent years, different kinds of depth fusion methods have emerged in different sub-tasks, such as stereo-ToF fusion~(\cite{Probablistic_Fusion,Reliable_Fusion,DL_Fusion}), stereo-stereo fusion~(\cite{Deep_stere_fusion}), Lidar-stereo fusion~(\cite{Probabilistic-Lidar-stereo,HP-Lidar-stereo-fusion}) and general depth fusion~(\cite{Sdf-GAN}). Additionally, deep-learning based methods perform much better than the rest. However, all of these algorithms are supervised and require much ground truth depth data to train. They suffer from two big problems: (1) Ground truth depth data is hard to get and expensive. (2) It is hard to generalize well with a limited amount of labeled data. As far as I know, the proposed algorithm in this paper is the first to develop a fully unsupervised depth fusion method, which solves the problems above and can fuse the depth inputs of different quality effectively without using ground truth depth data.

Unsupervised disparity fusion is hard because the algorithm requires to be able to produce an extremely accurate disparity map without any ground truth disparity data. The existing unsupervised strategy based on the left and right intensity consistency cannot guarantee a highly accurate disparity map. For example, Monodepth~\cite{Monodepth} treated the left-right intensity consistency error as a global metric in their cost function to obtain the disparity value but slight intensity changes in the images will influence the global estimation strongly. However, the left-right intensity consistency is just one of our local refinement metrics, which increases the global robustness and accuracy in turn. Previous work, such as our Sdf-MAN~\cite{Sdf-GAN}, achieved top disparity fusion performance but these algorithms need the ground truth disparity data to train. By combining the global disparity initialization with local disparity refinement, we show that our network can be trained without any ground truth depth data. 

In this paper, a fully unsupervised disparity fusion framework is proposed without the requirement of ground truth depth data. The initial disparity maps from the left view along with the auxiliary information (gradient, left \& right intensity image) are input into the refiner (a network similar to the generator in GAN~\cite{Standard_GAN} but without noise input) and the refiner is made to output the refined disparity map registered to the left view. The refined left disparity map and left intensity image are used to reconstruct the fake right intensity image. Finally, the fake and real right intensity images (from the right stereo vision camera) are fed into the discriminator (See Fig.~\ref{fig:Whole_diagram_UDFNet}). In the model, the refiner is trained to output a refined disparity value close to the weighted sum of the disparity inputs for global initialization (Equation~\ref{eq3_UDFNet}). Then, three refinement principles are adopted to refine the depth further. (1) The reconstructed intensity error between the fake and real right intensity image is minimized (Equation~\ref{eq1_UDFNet}). (2) The similarities between the fake and real right image in different receptive fields are maximized (Equation~\ref{eq4_UDFNet}). (3) The refined disparity map is smoothed based on the corresponding intensity image space (Equation~\ref{eq2_UDFNet}).

A novel and efficient network structure has been designed as well (See Figure~\ref{fig:Network_refiner_UDFNet} for refiner, Figure~\ref{fig:Network_discriminator_UDFNet} for discriminator). In the refiner, from the input layer to the bottleneck, the dense blocks and transition layers~\cite{Densenet} are used to increase local non-linearity to obtain more local detailed information. Long skip connections from previous layers to later layers are added to preserve the lost detail information after the bottleneck. The discriminator outputs the probability of the input image (real right image or reconstructed right image) being from the real distribution at different receptive field sizes (or different scales). Also, the dense blocks and transition layers have been adapted to increase the ability of the discriminator.

Section~\ref{Methodology_UDFNet} presents the pipeline of the proposed work, the mathematical model used in the design of the objective function for the network and the architecture for the refiner and discriminator. Section~\ref{Experiments_UDFNet} presents the experimental results. 

\textbf{Contributions:}
\begin{enumerate}
	\item An efficient unsupervised strategy by combining global disparity initialization and local refinement
	\item An indirect method using an adversarial network to force the disparity Markov Random Field of the refined disparity map to be close to the real
	\item An unsupervised end-to-end uncertainty-based pipeline to fuse any disparity input 
\end{enumerate}

\section{Methodology \label{Methodology_UDFNet}}
First a pipeline using a refiner network (similar to the generator in GAN~\cite{Standard_GAN} without noise input) is proposed to realize disparity fusion and then a mathematical model is built to design the cost function for the fully unsupervised network. Finally, the refiner and discriminator architectures are introduced. \\

\subsection{Fusion Pipeline}
\begin{figure}[h!]
	\subfloat[Refiner]{\includegraphics[width = 7cm]{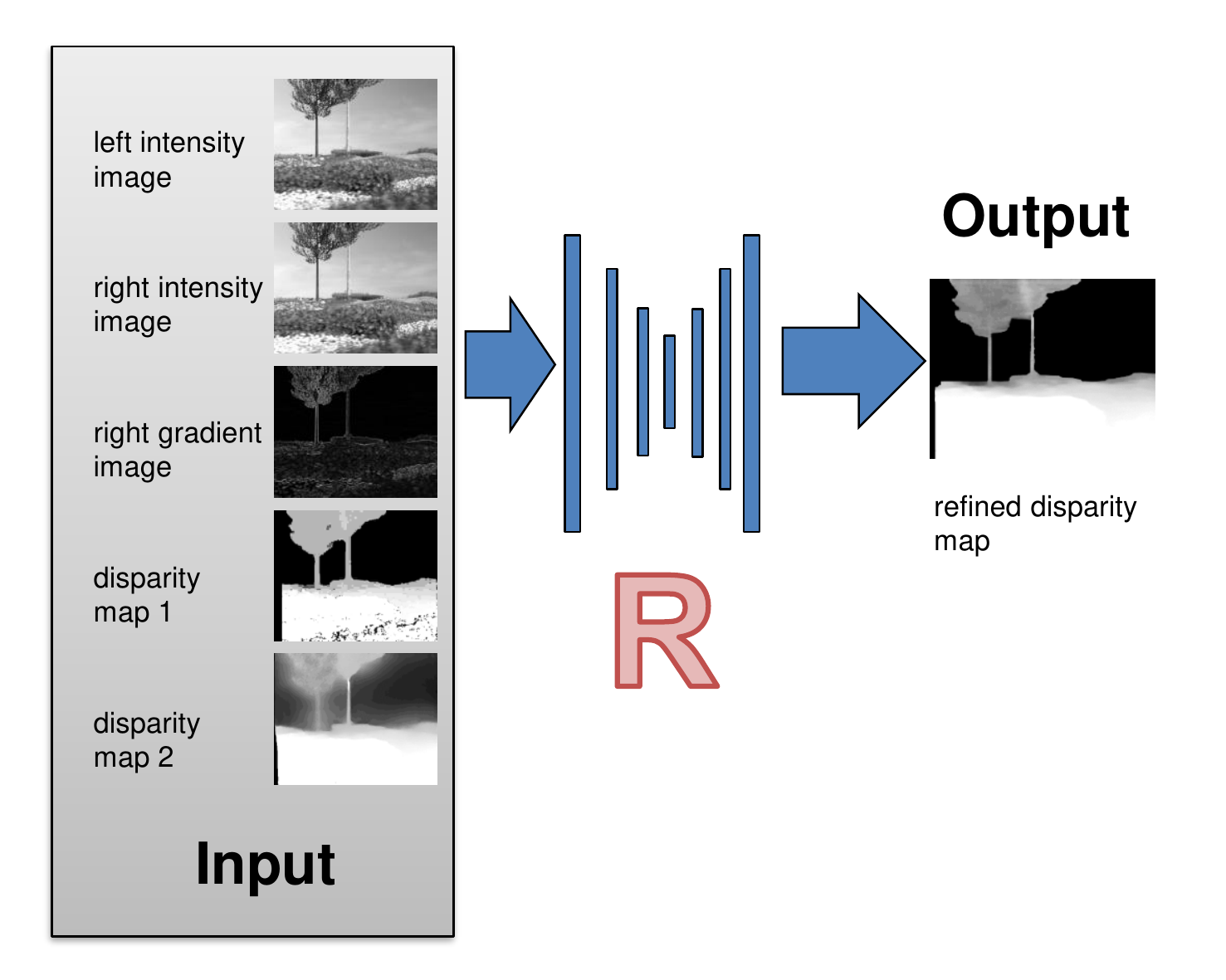}} 
	\hspace{0.5cm}
	\subfloat[Reconstruct right intensity image]{\includegraphics[width = 7cm]{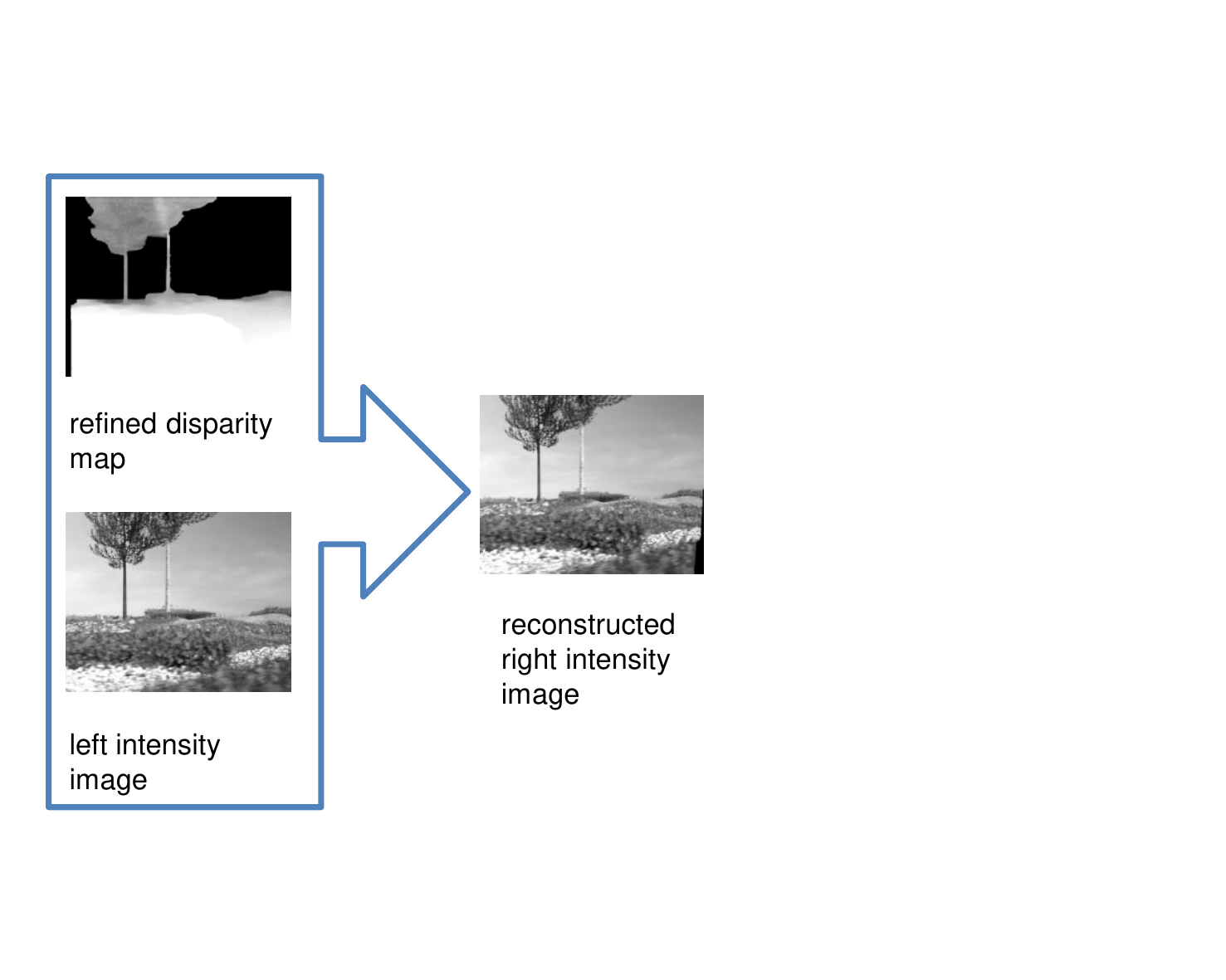}} \\
	
	\subfloat[Negative examples]{\includegraphics[width = 7cm]{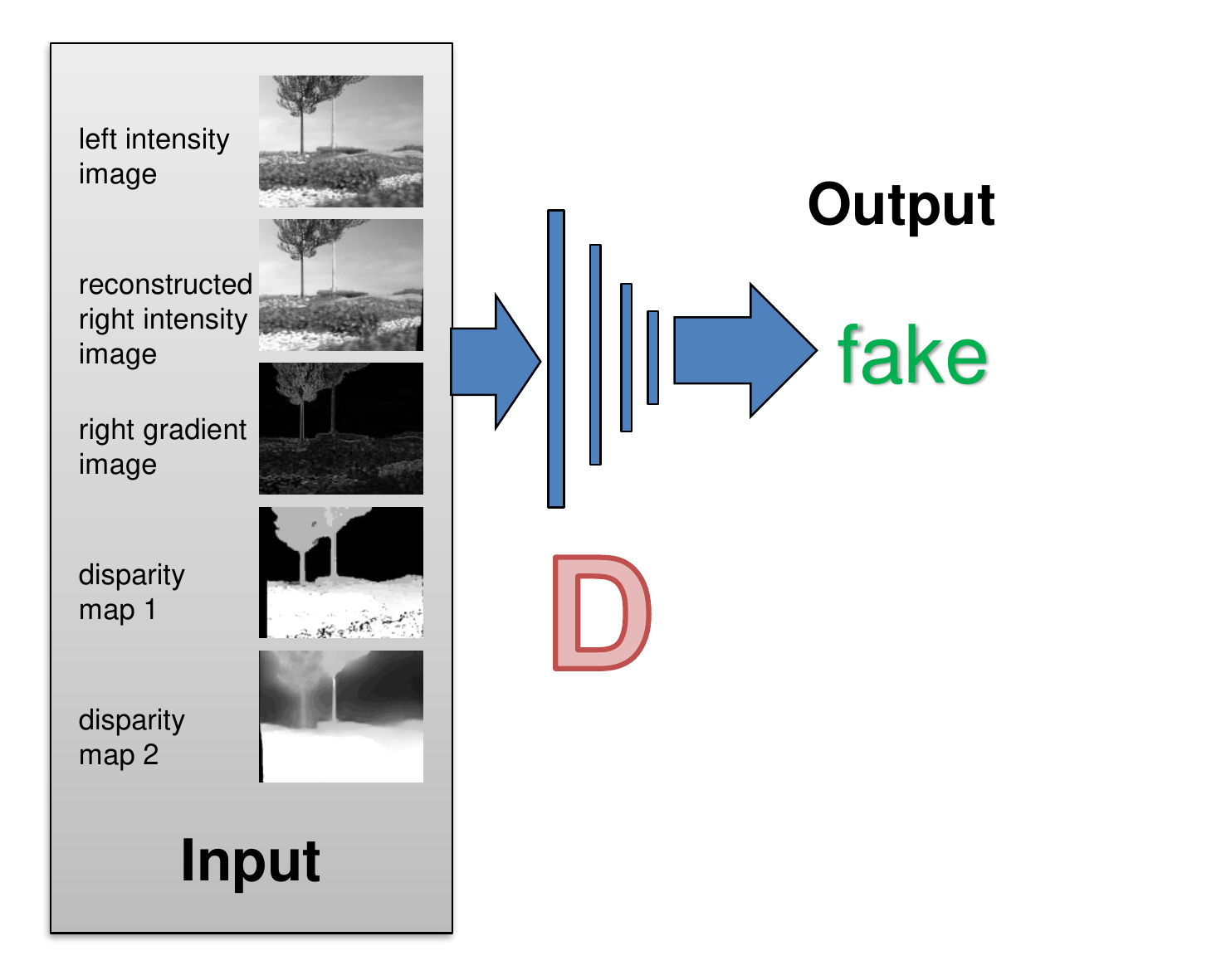}}
	\hspace{0.5cm}	 	
	\subfloat[Positive examples]{\includegraphics[width = 7cm]{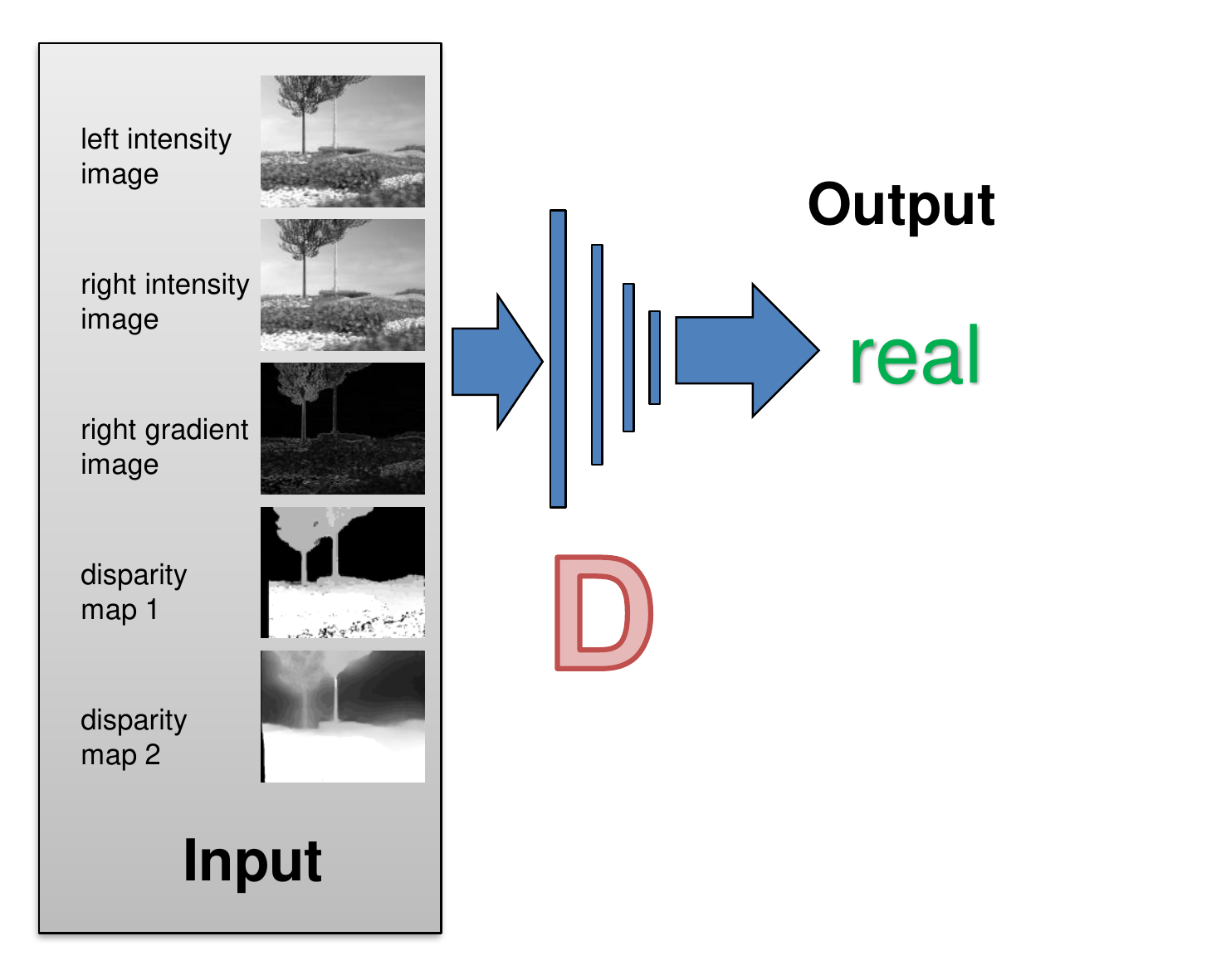}} \\
	
	\caption{\textcolor{black}{ (a): The inputs to the refiner (R) are the initial disparity maps (`disparity map 1', `disparity map 2') in the left view and auxiliary information (left intensity image, right intensity image, right gradient image). The gradient of the left view is calculated from the left intensity image directly. The refiner produces a refined disparity map in the left view. (b): The refined left disparity map and left intensity image are used to reconstruct the right intensity image. (c)(d): The input to discriminator (D) is the combination of the left intensity image, right gradient image, the initial disparity inputs and the reconstructed/real right image. The discriminator tries to discriminate whether the input is fake (reconstructed right image) in (c) or true (real right image) in (d). The images in each block come from the training process on a synthetic garden dataset.}}
	\label{fig:Whole_diagram_UDFNet}
\end{figure}

Similar to \cite{Sdf-GAN}, a refiner network has been used to map coarse disparity inputs to a real disparity distribution deterministically using multi-modal information (disparity, intensity, intensity gradient). Unlike ~\cite{Deep_stere_fusion, Sdf-GAN, DL_Fusion}, a fully unsupervised method is used to train the neural network without ground truth. Inspired by \cite{Monodepth}, the disparity output from the left view is treated as a hidden layer and used to reconstruct the intensity image of the right view. The whole process is shown in Figure~\ref{fig:Whole_diagram_UDFNet}.

\subsection{Objective Function}
Using the reconstructed intensity error as a metric usually cannot guarantee a highly precise disparity output (eg:~\cite{Monodepth}), especially in environments with similar color and repetitive patterns (eg: garden). Additionally, the fusion accuracy may decrease compared with highly accurate disparity inputs if only the reconstructed intensity error is used as the decision metric. Thus, a more complex mathematical model is proposed as the cost function for disparity fusion.

The main ideas for the mathematical model:

$\bullet$ The target is disparity fusion, whose accuracy should depend on the input disparity accuracy. The initial disparity maps should be used to provide the global initial value for the refined disparity map first. That is, the output of the refiner is encouraged to be similar to the input disparities (Equation~\ref{eq3_UDFNet}).

$\bullet$ The initialization based on Equation~\ref{eq3_UDFNet} can provide a coarse disparity map. The refinement will be realized by three local decision strategies discussed below together. The fake right intensity image is reconstructed from the left intensity image and disparity map. Thus, the accuracy of the refined disparity map can be assessed indirectly by comparing the reconstructed right image and real right image. The $L_1$ intensity error is designed based on the gradient in Equation~\ref{eq1_UDFNet} and the distance between the  Markov Random Field of the refined disparity map and real disparity distribution is described in Equation~\ref{eq4_UDFNet} indirectly. A disparity smoothness term is also designed to reduce the outliers and strong noise in Equation~\ref{eq2_UDFNet} using the gradient.

More specifically, the cost function has been designed as the following:

(1) Different from classical stereo vision algorithms without initial disparity inputs, disparity fusion is aimed at refinement of initial disparity inputs. Thus, a constraint is added that the output should be close to the weighted sum of the initial disparity inputs at each pixel. The output accuracy will increase with the accuracy enhancement of the initial disparity inputs. 

\begin{equation}
\label{eq3_UDFNet}
\mathcal{L}_{c}(G) = \mathop{{}\mathbb{E}}_{u \in \tilde{x}, u_s \in \bar{x}_s, \tilde{x} \sim P_G, s=1..Z}[w_{u_s} || \tilde{x}_u-\bar{x}_{u_s}||_1]
\end{equation} 

In Equation~\ref{eq3_UDFNet}, $\bar{x}_{u_s}$ is the  disparity value of a pixel $u_s$ of the $s^{th}$ initial disparity input (In Fig.~\ref{fig:Whole_diagram_UDFNet}, it is `disparity map 1' or `disparity map 2') corresponding to $u$ in the refined disparity output $\tilde{x}$ (In Fig.~\ref{fig:Whole_diagram_UDFNet}, it is `refined disparity map'). $P_G$ represents the distribution of the samples $\tilde{x}$ from the refiner.  $w_{u_s}$ is the confidence of the pixel $u_s$ from the $s^{th}$ initial disparity input. If no prior knowledge is available, $w_{u_s}=1/Z$ for all pixels. $Z$ is the number of initial disparity inputs.

(2) To encourage disparity estimates at edges to be more accurate, gradient information is integrated as a weight into the $L_1$ distance to make the disparity edges clearer. 

\begin{equation}
\label{eq1_UDFNet}
\mathcal{L}_{L_1}(G) = \mathop{{}\mathbb{E}}_{I_r \sim P_R, \tilde{I_r} \sim P_G' }[\exp(\alpha |\nabla (I_r)|) || I_r-\tilde{I_r}||_1]
\end{equation}  

In Equation~\ref{eq1_UDFNet}, $I_r$ is the real right intensity image from the right camera (In Fig.~\ref{fig:Whole_diagram_UDFNet}, it is `right intensity image'.) and $\tilde{I_r}$ is the reconstructed right intensity image from the refiner (In Fig.~\ref{fig:Whole_diagram_UDFNet}, it is `reconstructed right intensity image'). $\nabla (I_r)$ is the gradient of the gray image in the right view (In Fig.~\ref{fig:Whole_diagram_UDFNet}, it is `right gradient image' calculated from Sobel operator). $\alpha \geq 0$ weights the gradient value. $||\bullet ||_1$ is $L_1$ distance. $P_G'$ represents the distribution of the samples $\tilde{I_r}$ reconstructed from the left intensity image and corresponding refined disparity map. $P_R$ represents the distribution of the samples  $I_r$ from the right camera in the stereo vision setting. The goal is to encourage disparity estimates at edges (larger gradients) to be more accurate with less reconstructed intensity error.

(3) The right intensity image is reconstructed from the left intensity image using the refined disparity output. Unlike \cite{Sdf-GAN}, the reconstructed right intensity image and real right intensity image are input into the discriminator in this paper, which also gives indirect feedback about whether the refined disparity distribution is close to the ground truth. By making the discriminator output the probabilities at different receptive fields or scales (In Fig.~\ref{fig:Network_discriminator_UDFNet}, please refer to D1, D2,..,D5), the refiner will be forced to make the disparity distribution in the refined disparity map close to the real.

To alleviate training difficulties and avoid GAN mode collapse, the Improved WGAN loss function ~\cite{Improved-WGAN} is adopted. In Equation~\ref{eq4_UDFNet}, $D_i$ is the probability at the $i^{th}$ scale that the input image patch to the discriminator is from the real distribution at the $i^{th}$ scale.  $\lambda$ is the penalty coefficient. $\hat{I_r}$ is the random sample and $P_{\hat{I_r}}$ is its corresponding distribution. (For more details, please read \cite{Improved-WGAN}).

\begin{equation}
\label{eq4_UDFNet}
\begin{split}
\mathcal{L}_{wgan}(G,D_{i}) = \mathop{{}\mathbb{E}}_{\tilde{I_r} \sim P_G' }[D_{i}(\tilde{I_r})]-\mathop{{}\mathbb{E}}_{I_r \sim P_R}[D_{i}(I_r)] \\
+ \lambda \mathop{{}\mathbb{E}}_{\hat{I_r} \sim P_{\hat{I_r}} }[(||\nabla _{\hat{I_r}} D_{i}(\hat{I_r}) ||_2 - 1)^2]
\end{split} 
\end{equation}

(4) 
After the strategies above, the neural network may produce outliers (black holes) in the texture-less areas. In order to suppress the outliers and noise in the refined disparity map,  a gradient-based smoothness term is used to propagate more accurate disparity values to the areas with similar color by the assumption that the disparity in the neighborhood should be similar if the intensity is similar. However, this term tends to produce blurry edge in the refined disparity map.

\begin{equation}
\label{eq2_UDFNet}
\mathcal{L}_{sm}(G) = \mathop{{}\mathbb{E}}_{u \in \tilde{x}, v \in N(u), \tilde{x} \sim P_G }[\exp(\gamma-\beta | \nabla  (I_l)_{uv}|) || \tilde{x}_u-\tilde{x}_v||_1]
\end{equation} 

In Equation~\ref{eq2_UDFNet}, $\tilde{x}_u$ is the disparity of a pixel $u$ in the refined disparity map $\tilde{x}$ from the refiner. $\tilde{x}_v$ is the disparity value of a pixel $v$ in the neighborhood $N(u)$ of pixel $u$. $\nabla (I_l)_{uv}$ is the gradient in the left intensity image (the refined disparity map is produced on the left view) from pixel $u$ to pixel $v$. It is calculated from the left intensity image considering the diagonal, left and right directions. $\beta \geq 0$ and $\gamma \geq 0$ are responsible for how close the disparities are if the intensities in the neighborhood are similar.

(5) Finally, our object function for the fully unsupervised approach is:
\begin{equation}
\label{eq5_UDFNet}
\begin{split}
G^{*} = \arg \mathop{\min}\limits_{G} \mathop{\max}\limits_{D_{i}} [& \theta _{1}\mathcal{L}_{L_1}(G) + \theta _{2}\mathcal{L}_{sm}(G) + \theta _{3} \mathcal{L}_{c}(G)  \\
&- \theta _{4}\sum_{i=1}^{M}\mathcal{L}_{wgan}(G,D_{i})]
\end{split} 
\end{equation}

In Equation~\ref{eq5_UDFNet}, $M$ is the number of the scales (In Fig.~\ref{fig:Network_discriminator_UDFNet}, M=5, please refer to D1, D2, .., D5). $\theta _{1}$, $\theta _{2}$, $\theta _{3}$, $\theta _{4}$ are the weights for the different loss terms.

\subsection{Network architectures}
A fully convolutional neural network~\cite{FCN} is adopted and also the partial architectures from~\cite{DCGAN,Densenet,Sdf-GAN} are adapted here for the refiner and discriminator. The refiner and discriminator use dense blocks~\cite{Densenet} to increase local non-linearity.
Transition layers~\cite{Densenet} change the size of the feature maps to reduce the time and space complexity.
In each dense block and transition layer, modules of the form ReLu-BatchNorm-convolution are used. Two modules in the refiner and four modules in the discriminator in each dense block are used, where the filter size is 3$\times$3 and stride is 1. In each transition layer, only one module is used, where the filter size is 4$\times$4 and the stride is 2 (except that in Tran.3 of the discriminator the stride is 1).  

\begin{figure*}[h!]
	%\centering
	\includegraphics[width = 15cm]{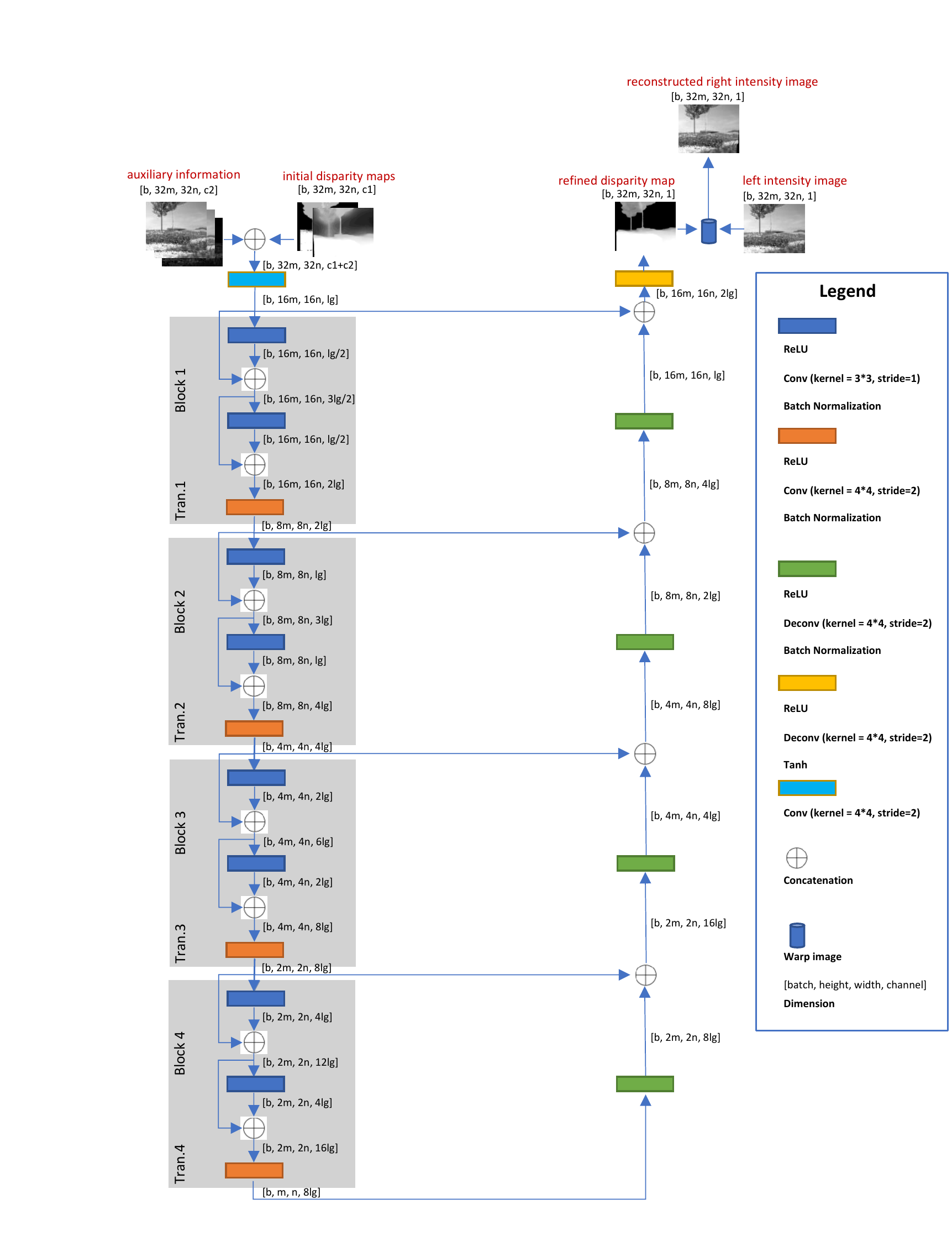}
	\caption{ This figure shows some important hyperparameters and the refiner architecture configuration. Please refer to Table~\ref{Network_setting_UDFNet} for the specific values in each experiment. }
	\label{fig:Network_refiner_UDFNet}
\end{figure*}

Figure~\ref{fig:Network_refiner_UDFNet} shows the main architecture of the refiner. c1 initial disparity inputs (the experiments below use $c1=2$ for 2 disparity maps) and c2 pieces of information (the experiments below use $c2=3$ for the left intensity image, the right intensity image and a gradient of the right intensity image)  are concatenated as input into the refiner. The batch size is b and resolution is 32m*32n. $lg$ is the number of the feature map channels after the first convolution. The refined disparity map is treated as a hidden layer in the network and used to reconstruct the intensity image in the right view. To reduce the computational complexity and increase the extraction ability of local details, each dense block contains only 2 internal layers. Additionally, the skip connections~\cite{Unet} from the previous layers to the latter layers preserve the local details in order not to lose information after the network bottleneck. During training, a dropout strategy has been added into the layers in the refiner after the bottleneck to avoid overfitting and the dropout part is cancelled during testing to produce a deterministic result. 

\begin{figure*}[h!]
	%\centering
	\includegraphics[width = 13cm]{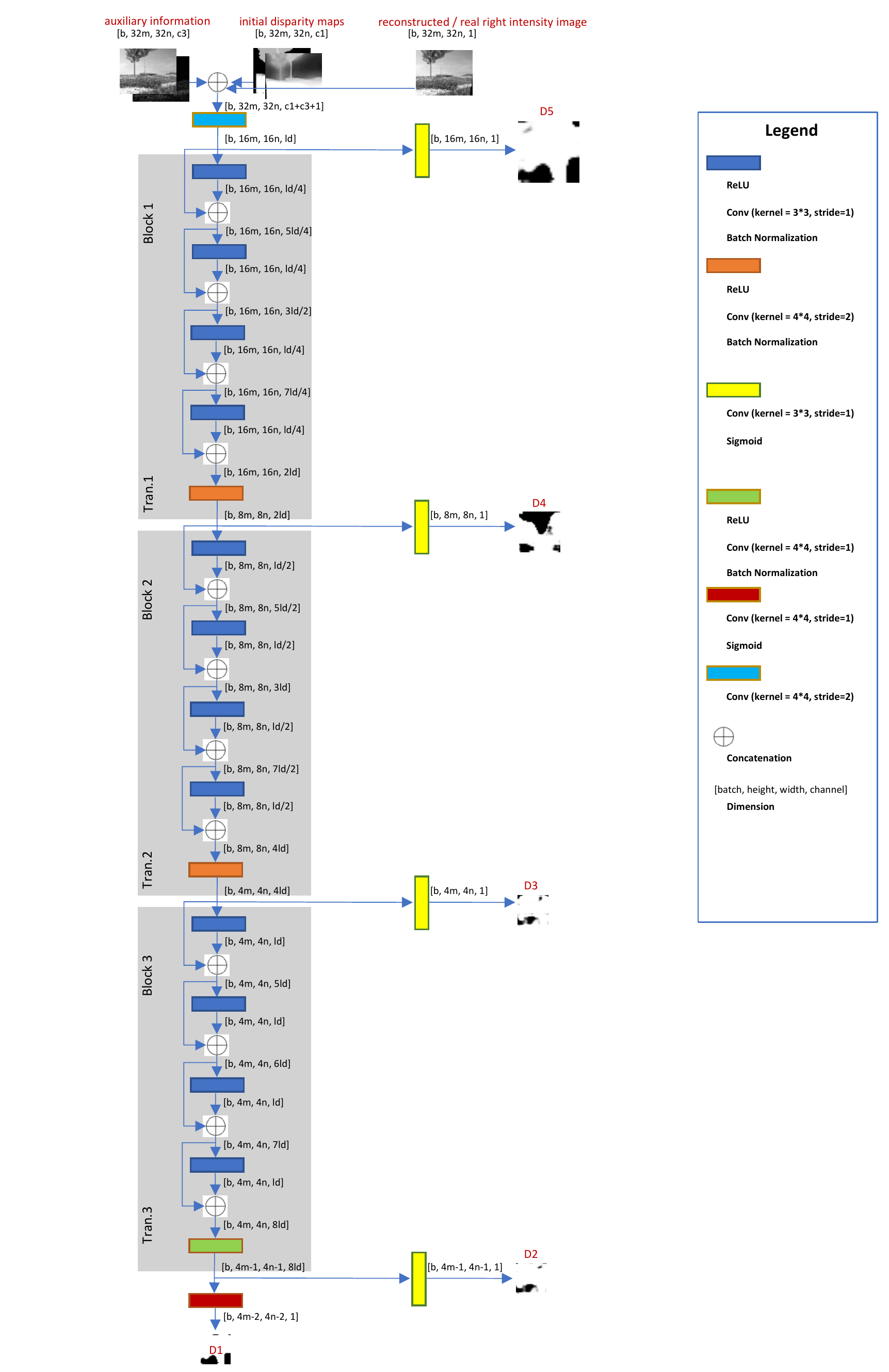}
	
	\caption{  This figure shows some important hyperparameters and the discriminator architecture configuration. Please refer to Table~\ref{Network_setting_UDFNet} for the specific values in each experiment. }
	\label{fig:Network_discriminator_UDFNet}
\end{figure*}

Figure~\ref{fig:Network_discriminator_UDFNet} is for the discriminator. The discriminator will only be used during training and abandoned during testing. Thus, the architecture of the discriminator will only influence the computational complexity during training. The initial disparity maps, information and real or reconstructed right images are concatenated and fed into the discriminator. Each dense block contains 4 internal layers. The sigmoid function (function tf.sigmoid in Tensorflow platform) outputs the probability map ($D_i, i=1,2..5$) that the input is real or fake at different scales to force the Markov Random Field of the refined disparity map to get closer to the real distribution at different receptive field sizes.

%%%%%%%%%%%%%%%%%%%%%%%%%%%%%%%%%%%%%%%%%%%%%%%%%%%%%%%%%%%%%%%%%%%%%%%%%%%%%%%

\section{Experiments \label{Experiments_UDFNet}}
The network is implemented using TensorFlow\cite{TensorFlow} and trained \& tested using an Intel Core i7-7820HK processor (quad-core, 8MB cache, up to 4.4GHZ) and Nvidia Geforce GTX 1080Ti. First, an ablation study with initial disparity inputs (~\cite{Dispnet, FPGA_stereo}) is conducted using a synthetic garden dataset to analyze the influence of each factor in the energy function. Secondly, a real test on the Kitti2015 dataset~\cite{Kitti2015} is done with two initial inputs (~\cite{PSMNet,SGM}). All the results show the proposed algorithm's superiority compared with the state-of-art or classical stereo vision algorithms (~\cite{PSMNet,FPGA_stereo,Dispnet,SGM}), and the state-of-the-art stereo-stereo fusion algorithms (~\cite{Sdf-GAN,Deep_stere_fusion}).    

In the following experiments, the inputs to the neural network were first padded to 32M * 32N (M, N are integers) using 0 and normalized to [-1, 1]. After that, the input was flipped vertically with a 50\% chance to double the number of training samples. Weights of all the neurons were initialized from a Gaussian distribution (standard deviation 0.02, mean 0). Each model is trained for 100 epochs on a synthetic garden dataset\footnote{It is not available to the public now.} and 500 epochs on Kitti2015, with a batch size 4 using Adam~\cite{Adam} with a momentum of 0.5. The learning rate is changed from 0.005 to 0.0001 gradually. The method in~\cite{Standard_GAN} is used to optimize the refiner and discriminator by alternating between one step on the discriminator and then one step on the refiner. The parameters  $\theta_{1}$, $\theta_{2}$, $\theta_{3}$, $\theta_{4}$ in Equation~\ref{eq5_UDFNet} were set to make those four terms contribute differently to the energy function in the training process. If the difference of two initial disparity values on the same pixel is small (\textless 0.3 pixels), a large value (0.99) is assigned to their confidence weight in Equation~\ref{eq3_UDFNet}. If not, they were set to a medium value (0.5). Besides the confidence estimation above, some other special empirical confidence estimation for some disparity inputs were adopted in the following experiments (For more details, see the corresponding experiments). Additionally, I didn't do any post-processing on the occlusion area and the other areas. The $L_1$ distance between the estimated value and ground truth is used as the error. The unit is pixel. For more details about the network settings and computational complexity, please see Table~\ref{Network_setting_UDFNet}. To highlight the real test, the network can do the disparity fusion (up to 384*1248 pixels) directly at 90 HZ without any cropping or downsampling.  

\begin{table*}[h!]
	\centering
	\caption{Computation Time and Initial Parameter Setting }
	\scalebox{.63}[.65]{\begin{tabular}{|p{0.8cm} | p{1.7cm}| p{1.8cm}| *{13}{p{0.9 cm}|}}  % repeats {c|} 6 times
			\hline
			& \multicolumn{15}{c|}{\textbf{Ablation Study} with Synthetic Garden Dataset }   \\ \hline
			\bf Para. &  Train time &  Test time &  b &  32m &  32n &  $c_1, c_2$, $c_3$ &   lg &   ld &   $\theta_{1}$ &   $\theta_{2}$ & $\theta_{3}$ &  $\theta_{4}$ &  $\alpha$ &  $\beta$ &  $\gamma$   \\ \hline 
			\bf Value & 0.22 (h/epoch)  & 0.010 (s/frame) & 4 & 384  & 480 & 2, 3, 2  & 32 & 32 & 10 & 0.1 & 0.001 & 1 & 3 & 650 & 5  \\ \hline
			& \multicolumn{15}{c|}{\textbf{Real Test } with Kitti2015 Dataset }   \\ \hline
			\bf Para. &  Train time &  Test time &  b &  32m &  32n &  $c_1, c_2$, $c_3$ &   lg &   ld &   $\theta_{1}$ &   $\theta_{2}$ & $\theta_{3}$ &  $\theta_{4}$ &  $\alpha$ &  $\beta$ &  $\gamma$   \\ \hline 
			\bf Value & 0.03 (h/epoch)  & 0.011 (s/frame) & 4 & 384  & 1248 & 2, 3, 2  & 10 & 10 & 1 & 20 to 1 & 0.0001 to~1K & 1 & 3 & 1 to 3000 & 5  \\ \hline

	\end{tabular} }
	\label{Network_setting_UDFNet}	
\end{table*}  

\begin{table*}[h!]
	\centering
	\caption{Ablation Study on Each Cue (Unit: Pixel) }
	\scalebox{.61}[.63]{\begin{tabular}{|p{2cm}| *{10}{p{1.7 cm}|}}  % repeats {c|} 6 times
			\hline
			\bf Experiment &  Input1~\cite{Dispnet} &  Input2~\cite{FPGA_stereo} & $\theta_{1}=0$  &  $\theta_{2}=0$ &  $\theta_{3}=0$ &  $\theta_{4}=0$ &  $\alpha=0$  &   $\beta=65$ &   $\gamma=1$ & Baseline   \\ \hline 
			\bf Error (pixel) & 4.67  & 8.52 & 3.53 & 3.94  & \bf 228.31 & 3.70  & 3.55 & 5.86 & 3.67 & \bf 3.45   \\ \hline
			\bf Tuning experience & overall accuracy  & overall accuracy & contrast influence  & smooth influence  & \bf initialized disparity & disparity MRF & accuracy at edge & smooth effect & smooth effect &    \\ \hline		
			
	\end{tabular} }
	\label{Ablation_study_UDFNet}	
\end{table*}

\subsection{Ablation Study}
\begin{figure*}[h!]
	\centering
	\subfloat[Scene]{\includegraphics[width = 7cm,height=5cm]{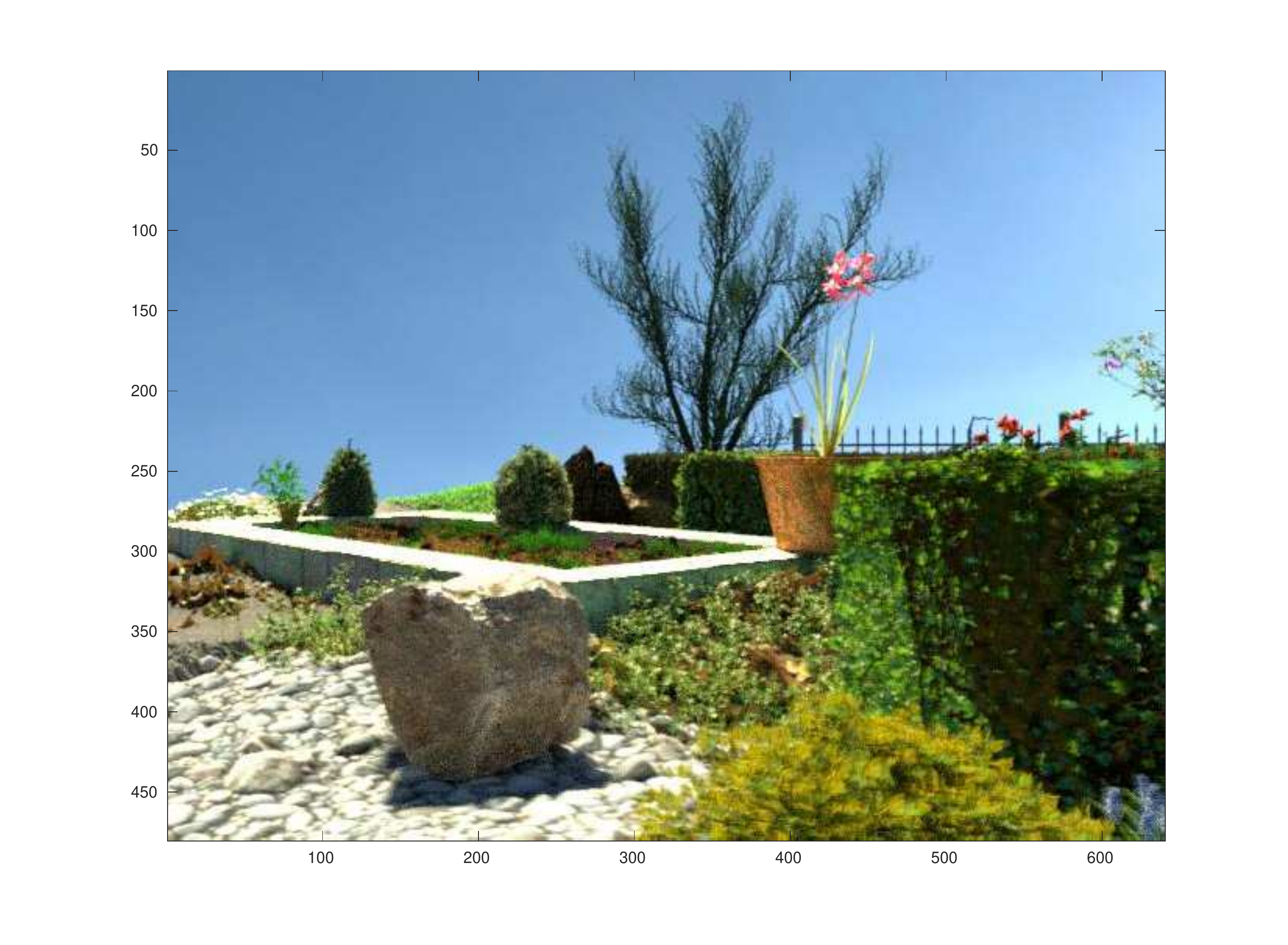}} 
	\vspace{0.001cm}
	\subfloat[Ground Truth]{\includegraphics[width = 7cm,height=5cm]{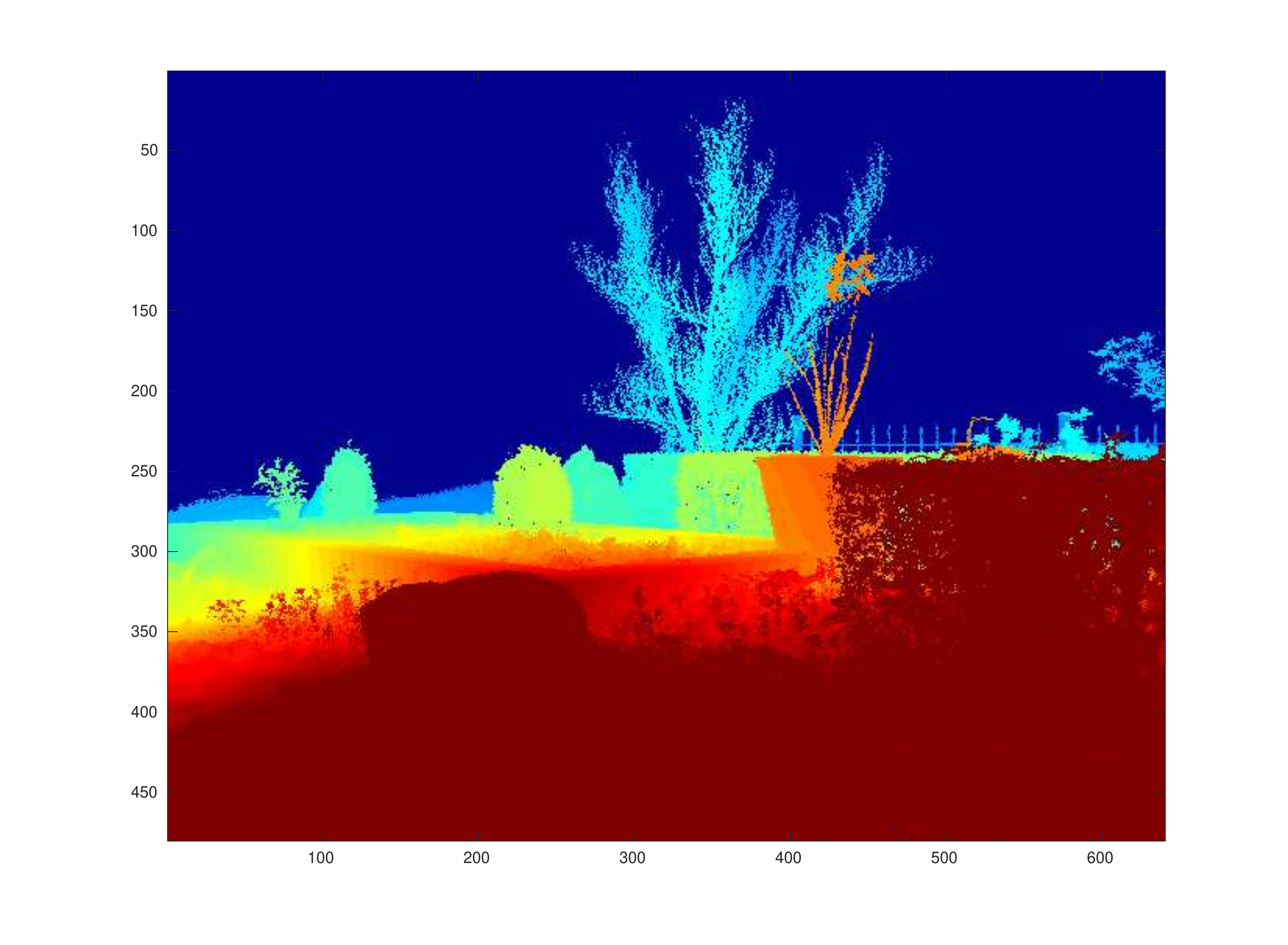}} 
	\vspace{0.001cm}	
	
	\subfloat[FPGA Stereo~\cite{FPGA_stereo}]{\includegraphics[width = 7cm,height=5cm]{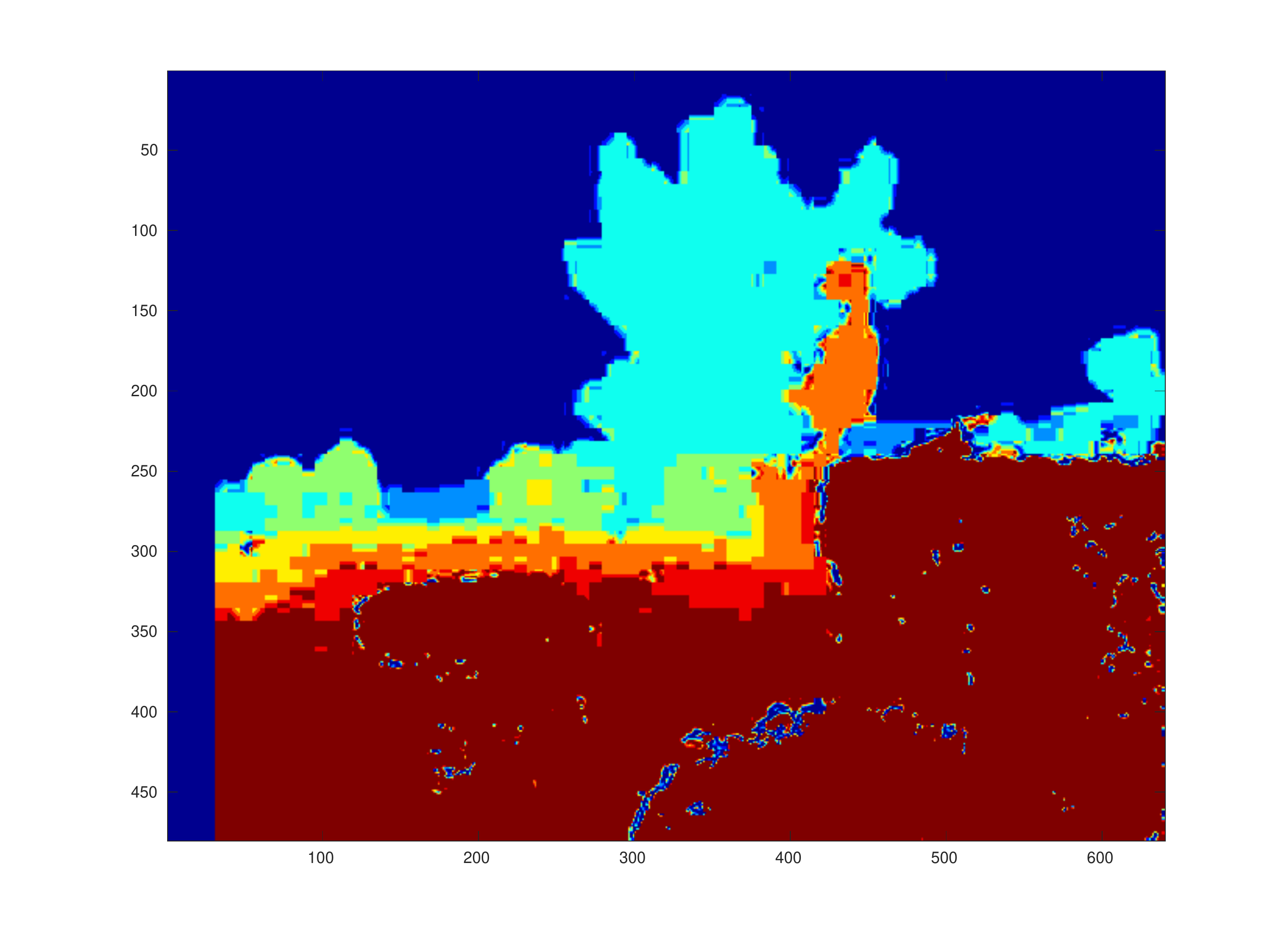}} 
	\vspace{0.001cm}
	\subfloat[Dispnet~\cite{Dispnet}]{\includegraphics[width = 7cm,height=5cm]{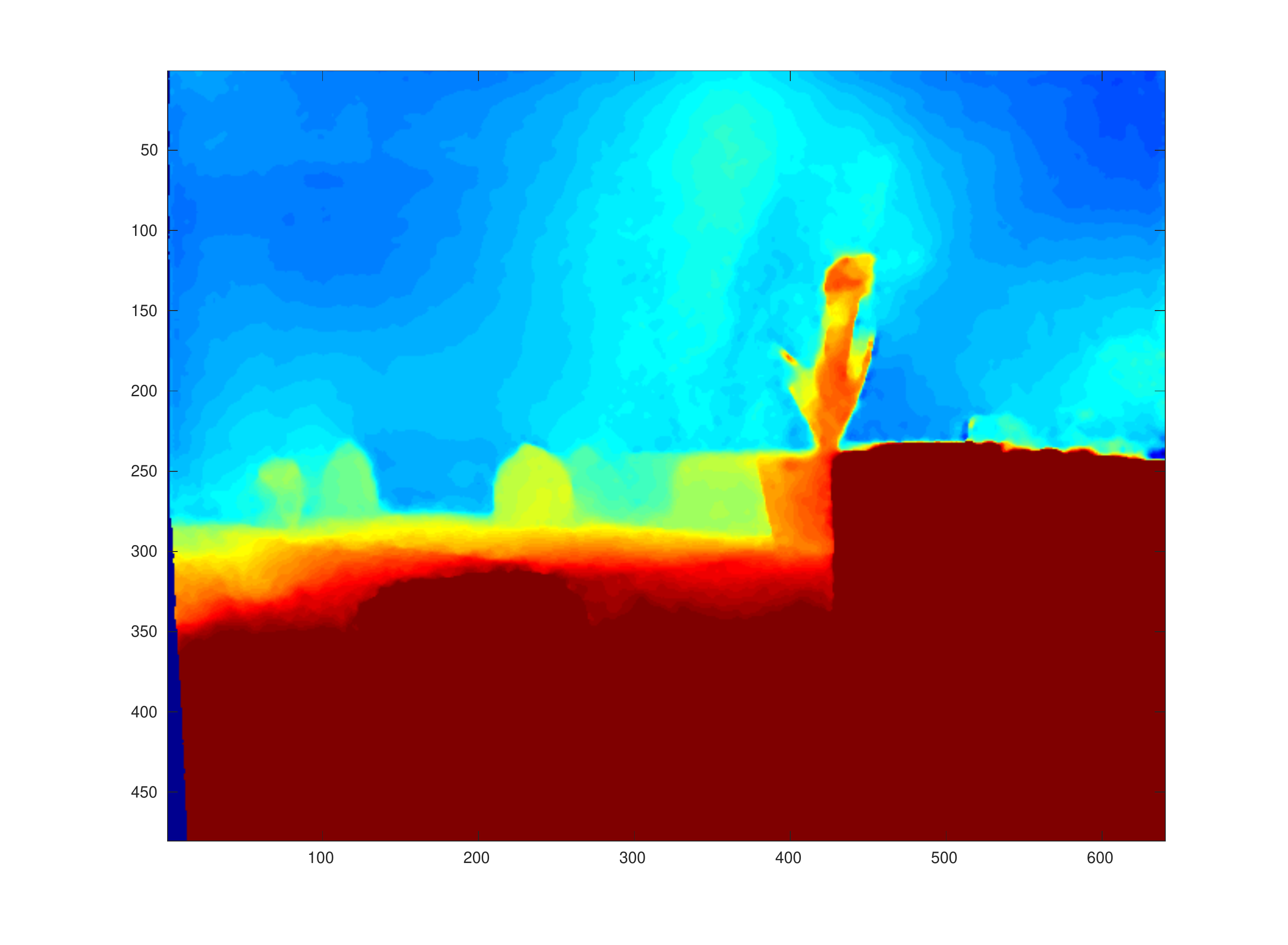}} 
	\vspace{0.001cm} 
	
	\subfloat[Fused Disparity]{\includegraphics[width = 7cm,height=5cm]{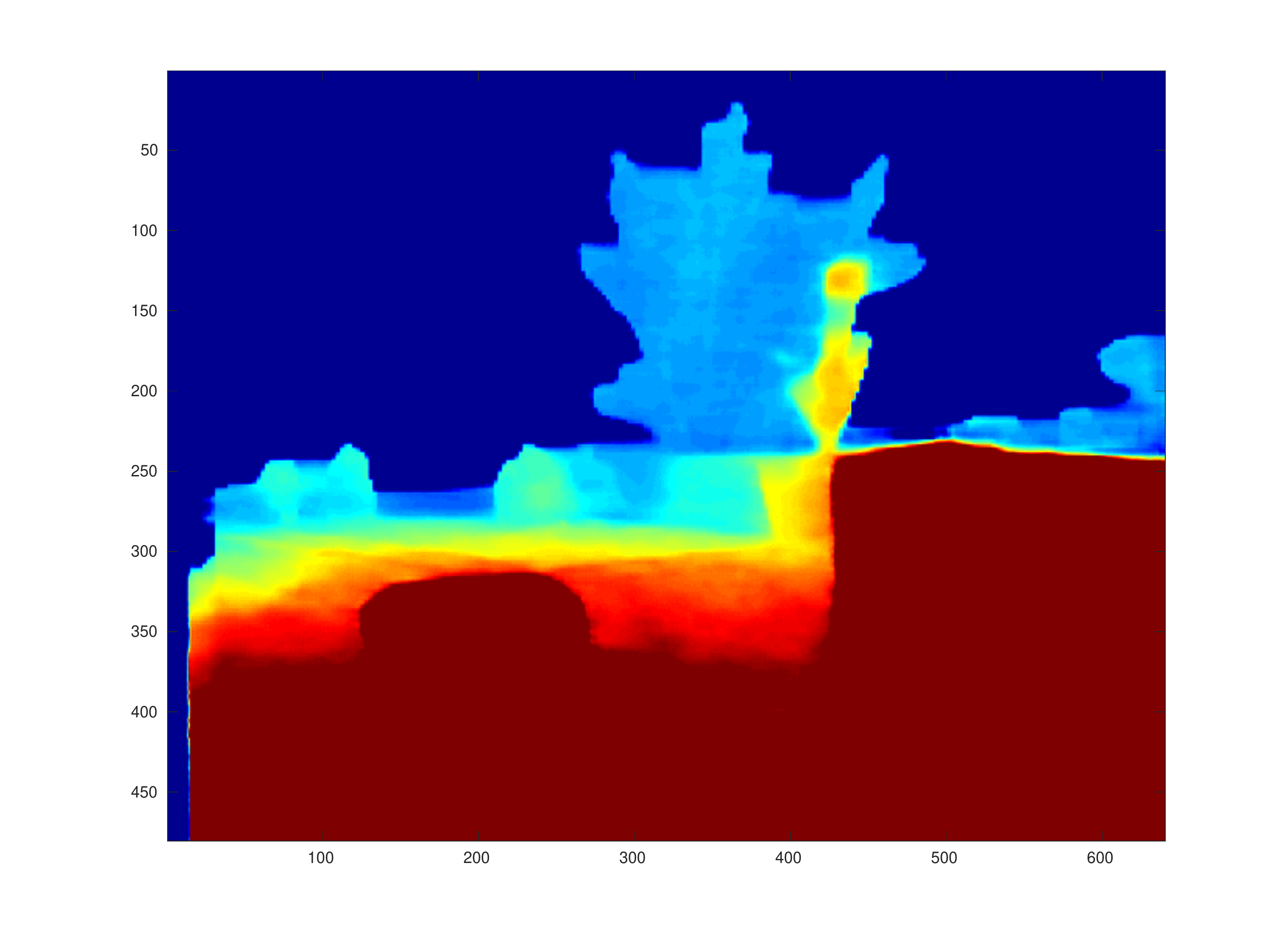}} 
	\vspace{0.001cm}

	\caption{Two initial disparity inputs (c)(d) were fused to get a refined disparity map (e) using our baseline method on the synthetic garden dataset. (b) is the ground truth and (a) is the corresponding scene. Many, but not all, pixels from the fused result are closer to the ground truth than the original inputs. Tip: Zooming in the electronic version gives a better view.}
	\label{fig:Ablation_study_UDFNet}
\end{figure*}

The synthetic garden dataset  contains 4600 training samples and 421 test samples under outdoor environments. Each sample has one pair of rectified stereo images and dense ground truth with resolution 480*640 (height * width) pixels. The reason using a synthetic dataset is that the real dataset (eg: Kitti2015) does not have dense ground truth, which will influence the evaluation of the network. Dispnet~\cite{Dispnet} and FPGA-stereo~\cite{FPGA_stereo} were used as inputs. The authors of ~\cite{Dispnet,FPGA_stereo} helped us get the best performance on the dataset as the input to the network. Besides the confidence estimation strategy above, another special rule to Dispnet was added because the disparity map from Dispnet is very inaccurate in remote areas. That is, the pixels whose disparity is less than 4 (pixels) have a small confidence (0.1) and, otherwise, have a big confidence (0.9) because it is more accurate for close scenes. The default settings for the baseline network is shown in Table~\ref{Network_setting_UDFNet}. In the ablation study, one of the following factors ($\theta_{1}$, $\theta_{2}$, $\theta_{3}$, $\theta_{4}$, $\alpha$, $\beta$, $\gamma$) in our energy function is changed to see the influence of each cue. The performance results are listed in Table~\ref{Ablation_study_UDFNet}. One example is Figure~\ref{fig:Ablation_study_UDFNet}.  As said in the introduction, the disparity constraint term $\theta_{3}$ (Equation \ref{eq3_UDFNet}) encourages the network to produce disparities close to the initial disparity inputs. It is the most important factor (228.31 pixels errors) because the other factors do not provide an accurate global initialization but mainly refine the disparity value in the local area using pixel-pixel intensity ($\theta_{1}$,$\alpha$), smoothness ($\theta_{2}$, $\gamma$, $\beta$) and Markov Random Field ($\theta_{4}$) . When tuning the network, the experience also followed the theory in the methodology section (Table~\ref{Ablation_study_UDFNet}).

To explore the influence of the input accuracy and the number of initial inputs, one or two initial disparity maps with different noise levels were input into the networks. The disparity inputs were got by adding different levels of Gaussian noise $N(0,\sigma^2)$  to the normalized ground truth (ground truth divided by 480). The confidence for each pixel in the experiments is proportional to the noise value under the assumption that accurate confidence estimation in the future can be obtained. The fusion accuracy (Table~\ref{Ablation_Study_2_UDFNet}) increases with the number and accuracy of the disparity inputs. When there is only one input with $\sigma=0.002$, the proposed method still can refine the input disparity (0.77 \textless 1 pixel ) because there are accurate confidence estimates. The proposed method sometimes fails to refine only one input disparity map when there are no confidence estimates (all confidences are 100\%) and the mean error (\textless 1 pixel) is too small. Otherwise, it works robustly and effectively. 

\begin{table}[h!]
	\centering
	\caption{Average error (pixel) on Synthetic Garden } \label{Ablation_Study_2_UDFNet}	
	\begin{tabular}{|p{1.8cm}|*{5}{p{1 cm}|}}  % repeats {c|} 6 times
		\hline
		& \multicolumn{2}{c|}{\bf Fuse one input } & \multicolumn{3}{c|}{\bf Fuse two inputs } \\ \hline
		\bf Noise std. &  \bf input1 &  \bf fused &  \bf input1 &  \bf input2 &  \bf fused  \\ \hline 
		$\sigma=0.002$ & 0.77  & 0.69 & 0.77 & 0.77  & 0.64  \\ \hline
		$\sigma=0.004$ & 1.53 & 1.13 & 1.53 & 1.53 & 0.71 \\ \hline
		$\sigma=0.008$ & 3.06 & 1.58 & 3.06 & 3.06 & 0.80  \\ \hline	
		$\sigma=0.016$ & 6.12  & 3.05 & 6.12 & 6.12 & 1.33  \\ \hline			
	\end{tabular}	
\end{table}

\subsection{Real Data}
\subsubsection{Stereo-stereo Fusion}

The network was tested on the real Kitti2015 dataset, which used a Velodyne HDL-64E Lidar scanner to get the sparse ground truth and a 1242*375 resolution stereo camera to get stereo image pairs. The training dataset contains 400 unlabeled and labeled samples in all. There are another 400 samples in the test dataset. 50 samples from `000000\_10.png' to `000049\_10.png' 
in the Kitti2015 training dataset were used as our test dataset. 50 samples from `000050\_50.png' to `000099\_10.png' 
in the Kitti2015 training dataset were used as our validation dataset. The rest 700 samples were used as our training set. By flipping the training samples vertically, it doubled the number of training samples. The state-of-art stereo vision algorithm PSMNet~\cite{PSMNet} was used as one of our inputs. Their released pre-trained model\footnote{PSMNet~\cite{PSMNet}: \url{https://github.com/JiaRenChang/PSMNet}} on the Kitti2015 dataset was used to get the disparity maps. A traditional stereo vision algorithm SGM~\cite{SGM} was used as the second input to the network. Because the sparsity of SGM is not so important, its parameters were set to produce more reliable disparity maps. Thus, big confidence values (0.8) were assigned to its valid pixels and 0 to its invalid pixels' confidences. More specifically, the implementation (`disparity' function) from Matlab2016b was used. The relevant parameters are: 'DisparityRange' [0, 160], `BlockSize' 5, `ContrastThreshold' 0.99, `UniquenessThreshold' 70, `DistanceThreshold' 2. The settings of the neural network are shown in Table~\ref{Network_setting_UDFNet}. The proposed algorithm was compared with the state-of-the-art technique~\cite{Sdf-GAN,Deep_stere_fusion} in stereo-stereo fusion and also stereo vision inputs~\cite{PSMNet,SGM}.  The proposed method was compared with the supervised method in Sdf-MAN~\cite{Sdf-GAN}. The supervised method in Sdf-MAN was trained on our synthetic garden dataset first and then fine-tuned on the Kitti2015. 150 labeled samples from `00050\_10.png' to `000199\_10.png' 
in the initial training dataset were used for Sdf-MAN fine-tuning. Compared~with SGM~\cite{SGM} (0.78 pixels)~({This is a more accurate disparity but is calculated only using more reliable pixels. On~average only 40\% of the ground truth pixels are used. If all the valid ground truth are used to calculate its error, it~is 22.13 pixels}), the~fused results are much more dense and accurate. The performance of the proposed algorithm (0.83 pixels) (See Table~\ref{Real Test_UDFNet}) is  better than Sdf-MAN (1.17 pixels). The reason is because Sdf-MAN can't generalize in the real environment well. However, the proposed algorithm is not affected by such problems because the unsupervised method can use the unlabeled data directly.

\begin{table}[h!]
	\centering
	\caption{Average error (pixel) on Kitti2015 } \label{Real Test_UDFNet}	
	\begin{tabular}{|p{1.9cm}| p{1.9cm}| *{3}{p{2.7 cm}|}}  % repeats {c|} 6 times
		\hline
		\multicolumn{2}{|c|}{\bf Source Error } & \multicolumn{3}{c|}{\bf Fused Algorithm Error } \\ \hline		
		SGM \cite{SGM} &  PSMNet~\cite{PSMNet} &  DSF~\cite{Deep_stere_fusion} &  Sdf-MAN~\cite{Sdf-GAN} &   Ours  \\ \hline 
		0.78  & 1.22 & 1.20 & 1.17  & 0.83  \\ \hline
		
	\end{tabular}	
\end{table}   

\begin{figure*}[h!]
	
	\centering
	\subfloat[Ground Truth]{\includegraphics[height=3.5cm]{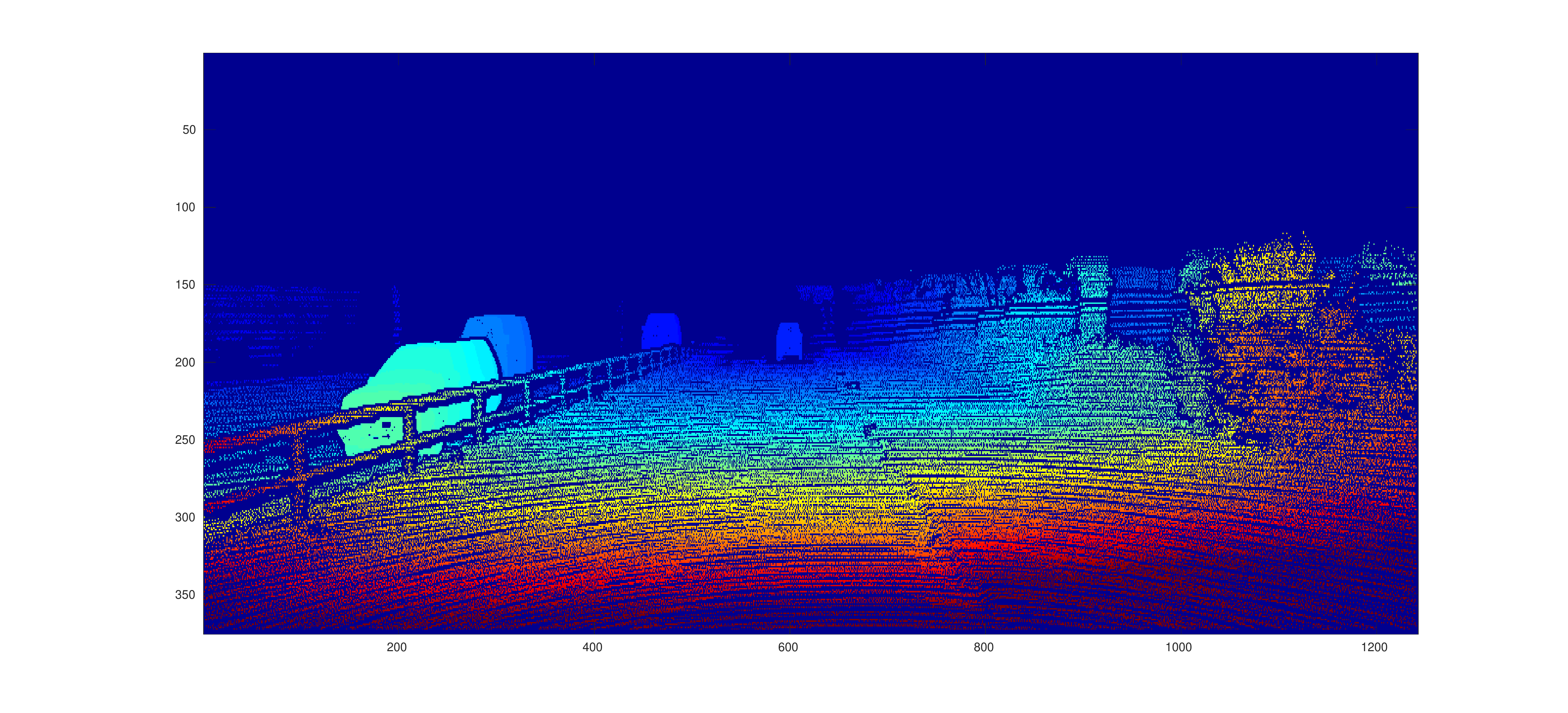}} 
	\subfloat[Scene]{\includegraphics[height=3.5cm]{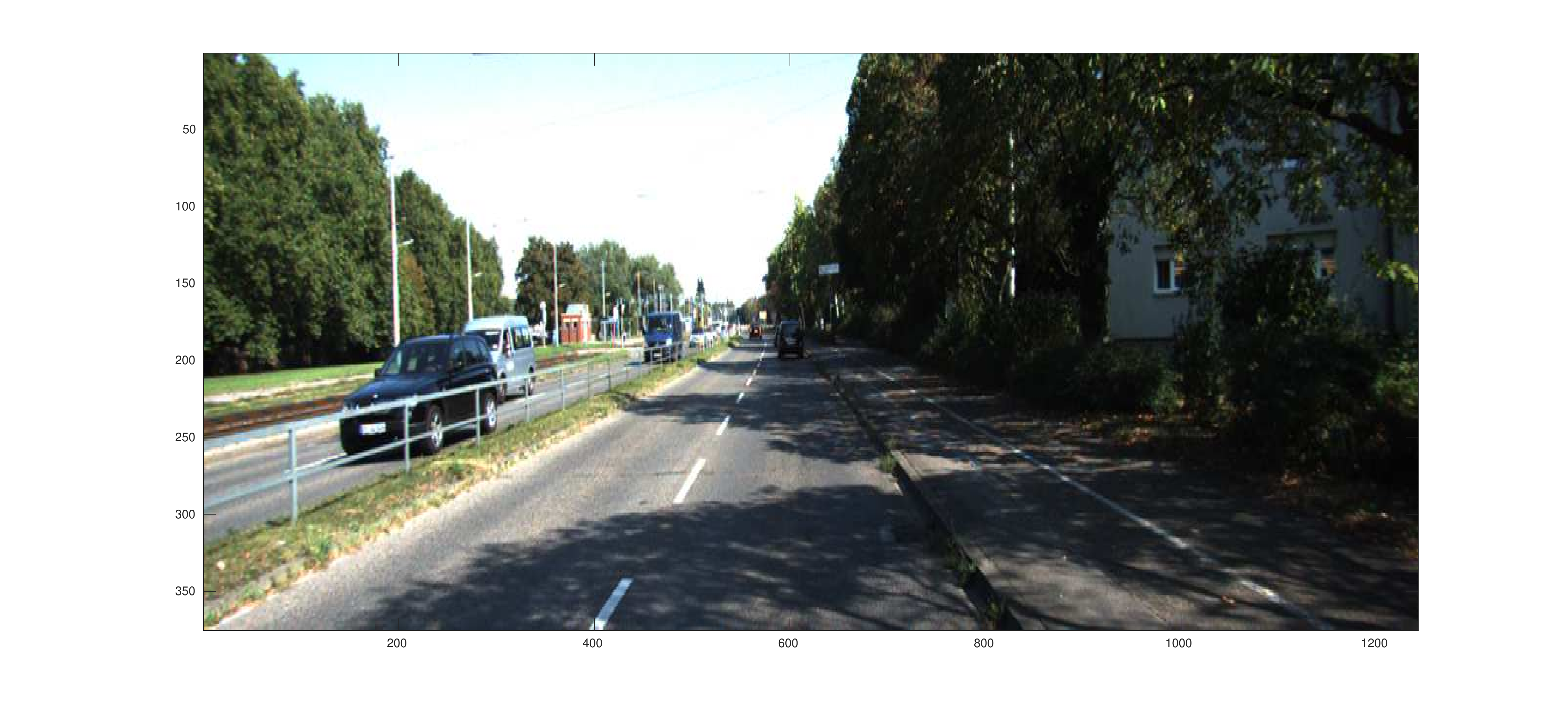}}\\
	
	\subfloat[Input Disparity 1: SGM~\cite{SGM}]{\includegraphics[height=3.5cm]{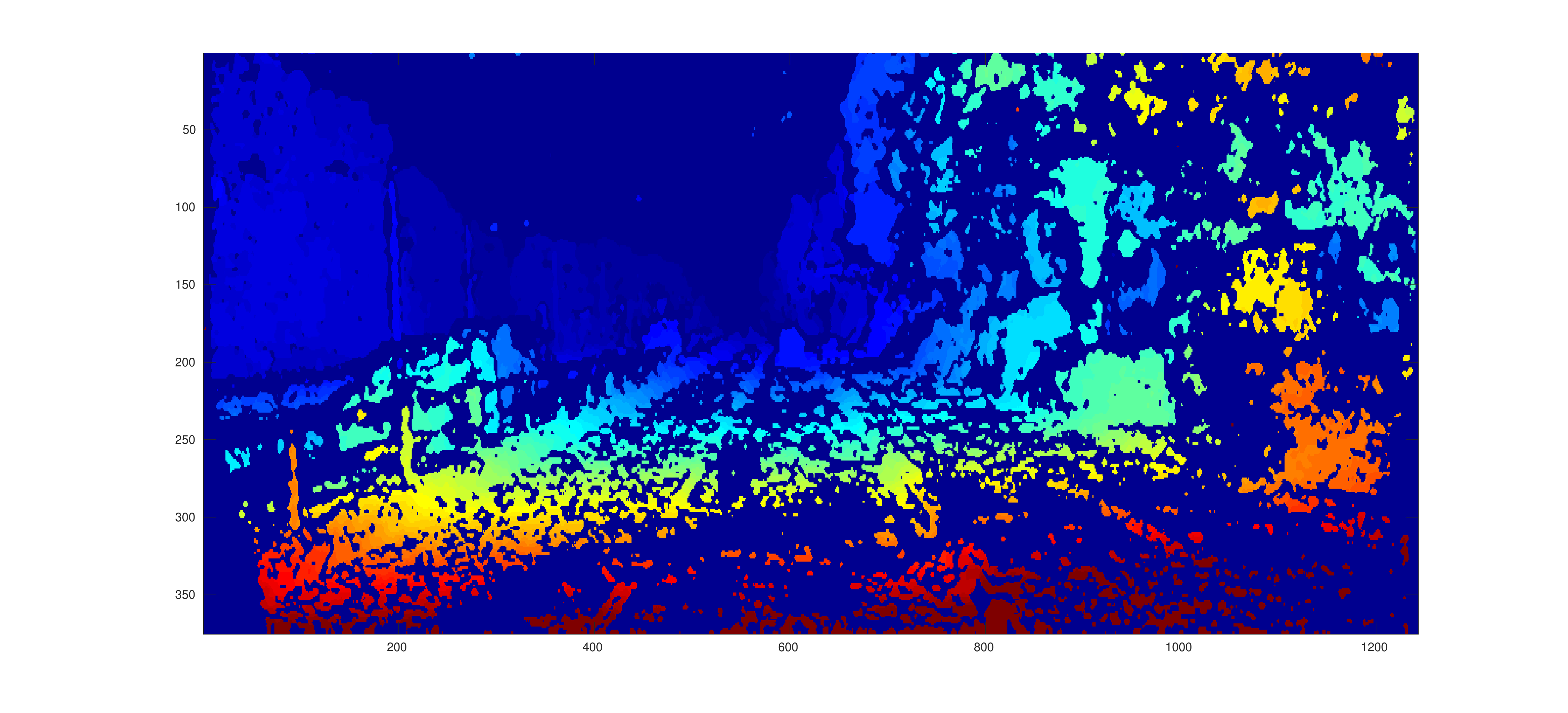}}
	\subfloat[Input Disparity 1 Error: SGM~\cite{SGM}]{\includegraphics[height=3.5cm]{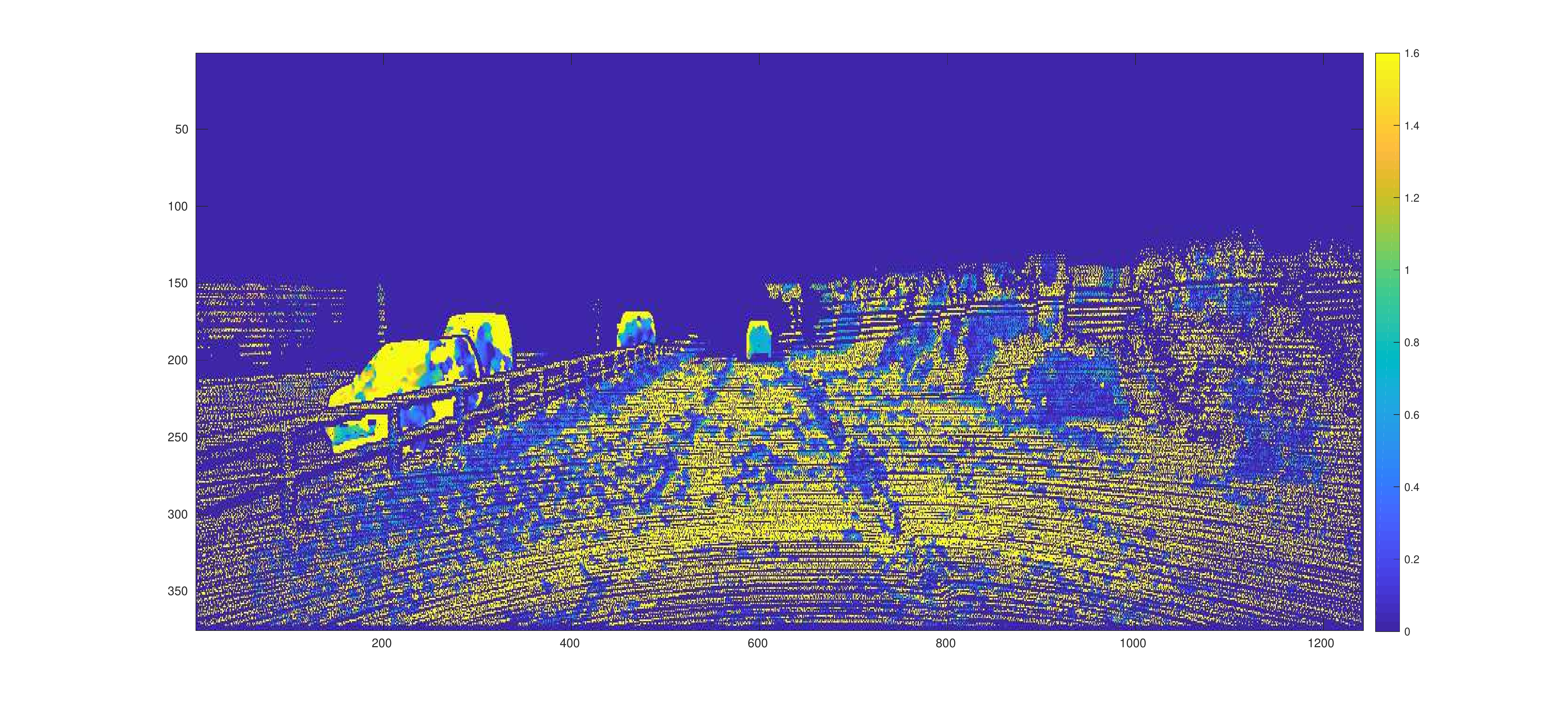}} \\
	
	\subfloat[Input Disparity 2: PSMNet~\cite{PSMNet}]{\includegraphics[height=3.5cm]{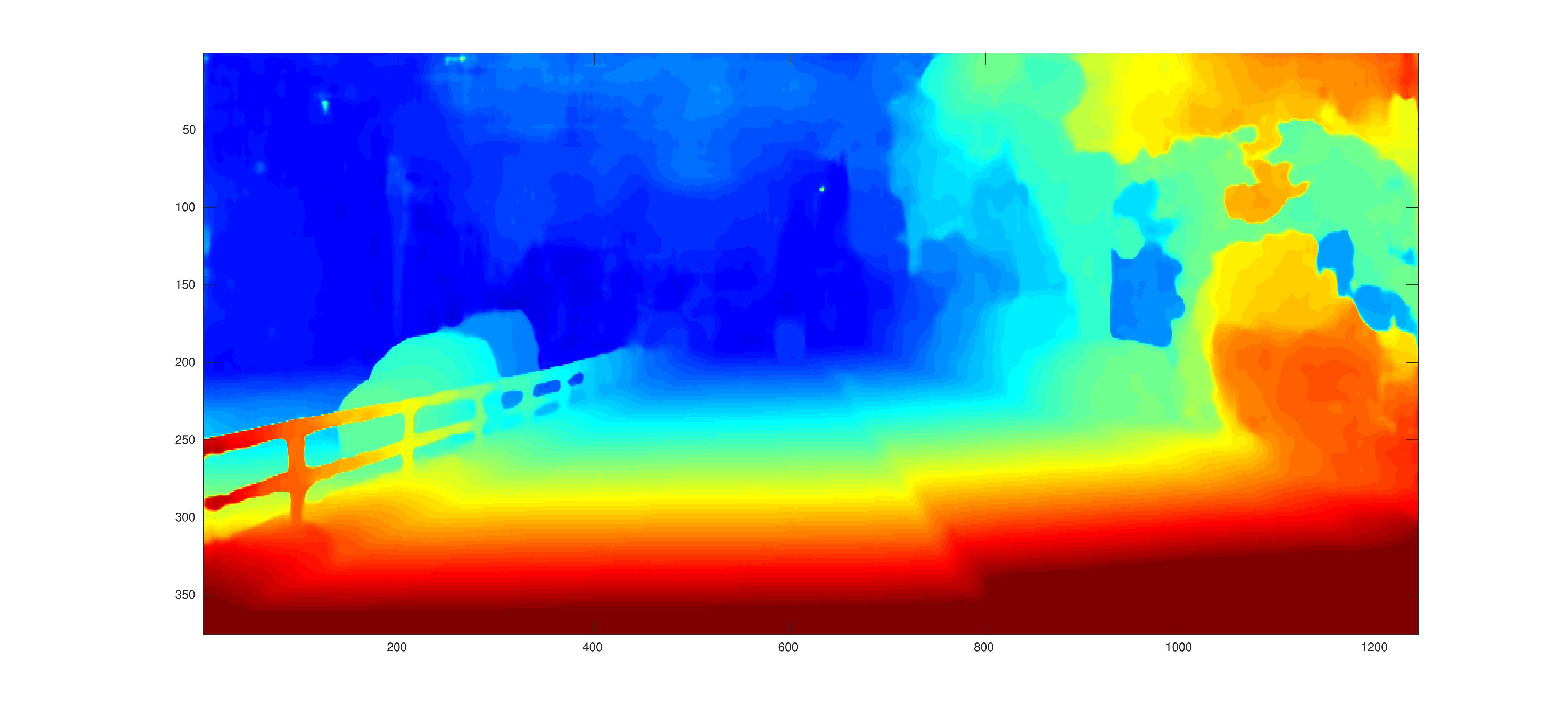}}
	\subfloat[Input Disparity 2 Error: PSMNet~\cite{PSMNet}]{\includegraphics[height=3.5cm]{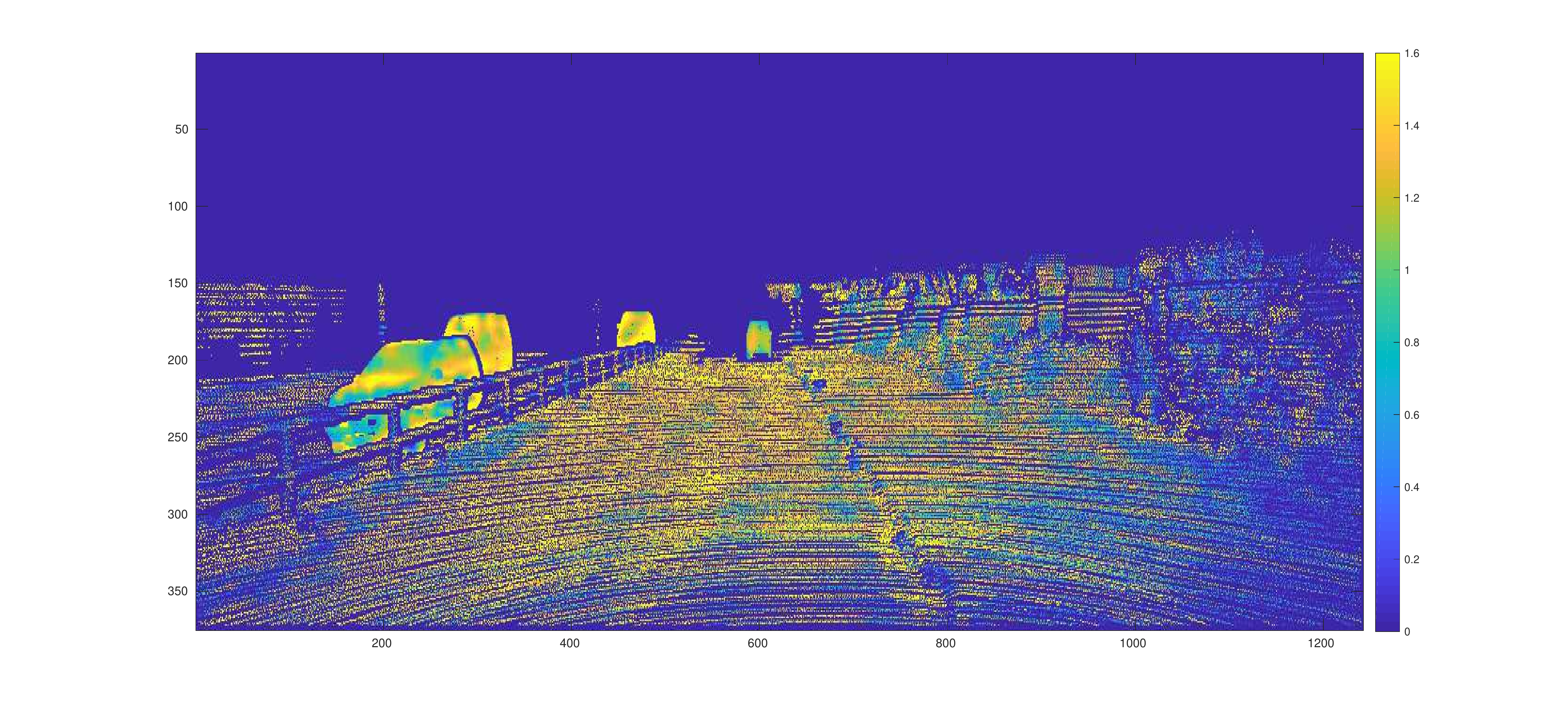}} \\
	
	\subfloat[Refined Disparity]{\includegraphics[height=3.5cm]{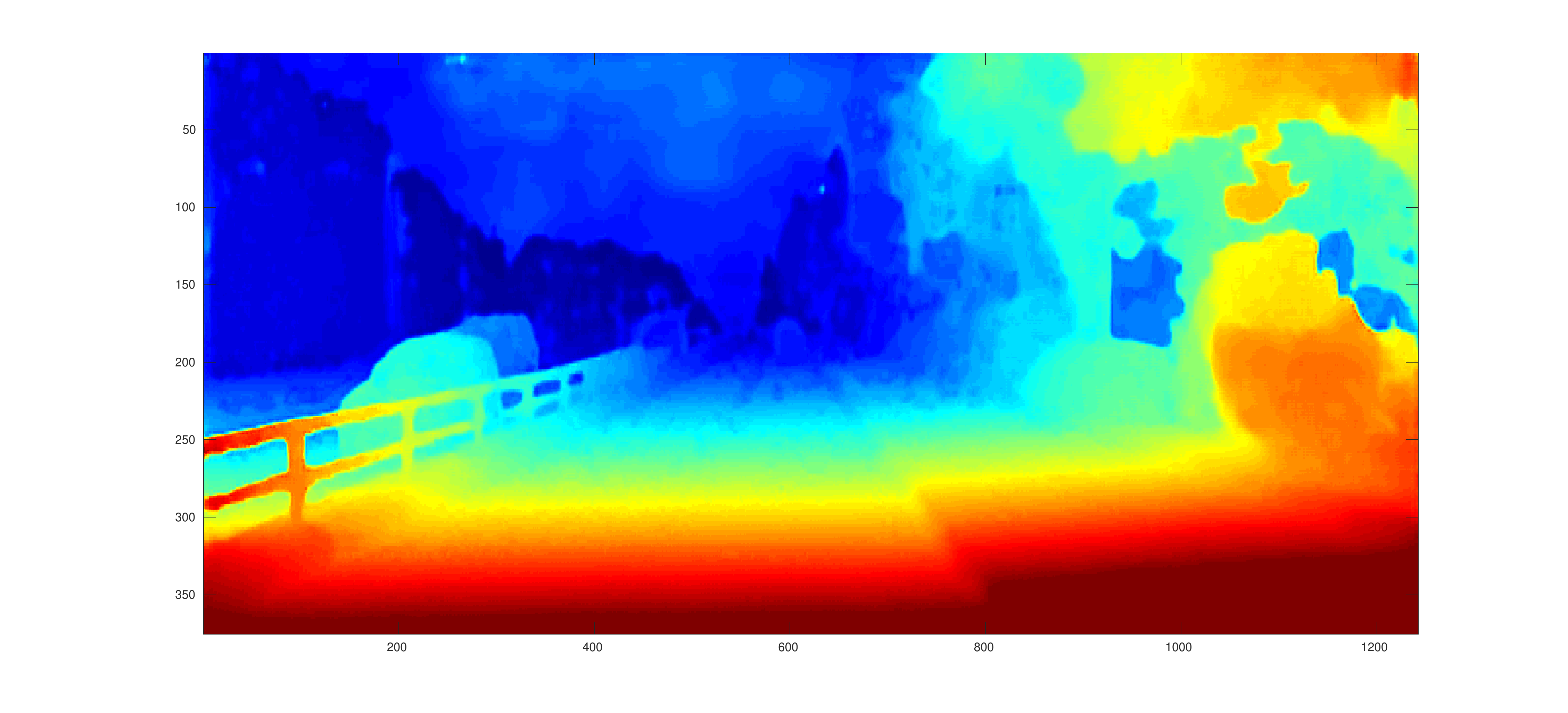}} 
	\subfloat[Refined Disparity Error]{\includegraphics[height=3.5cm]{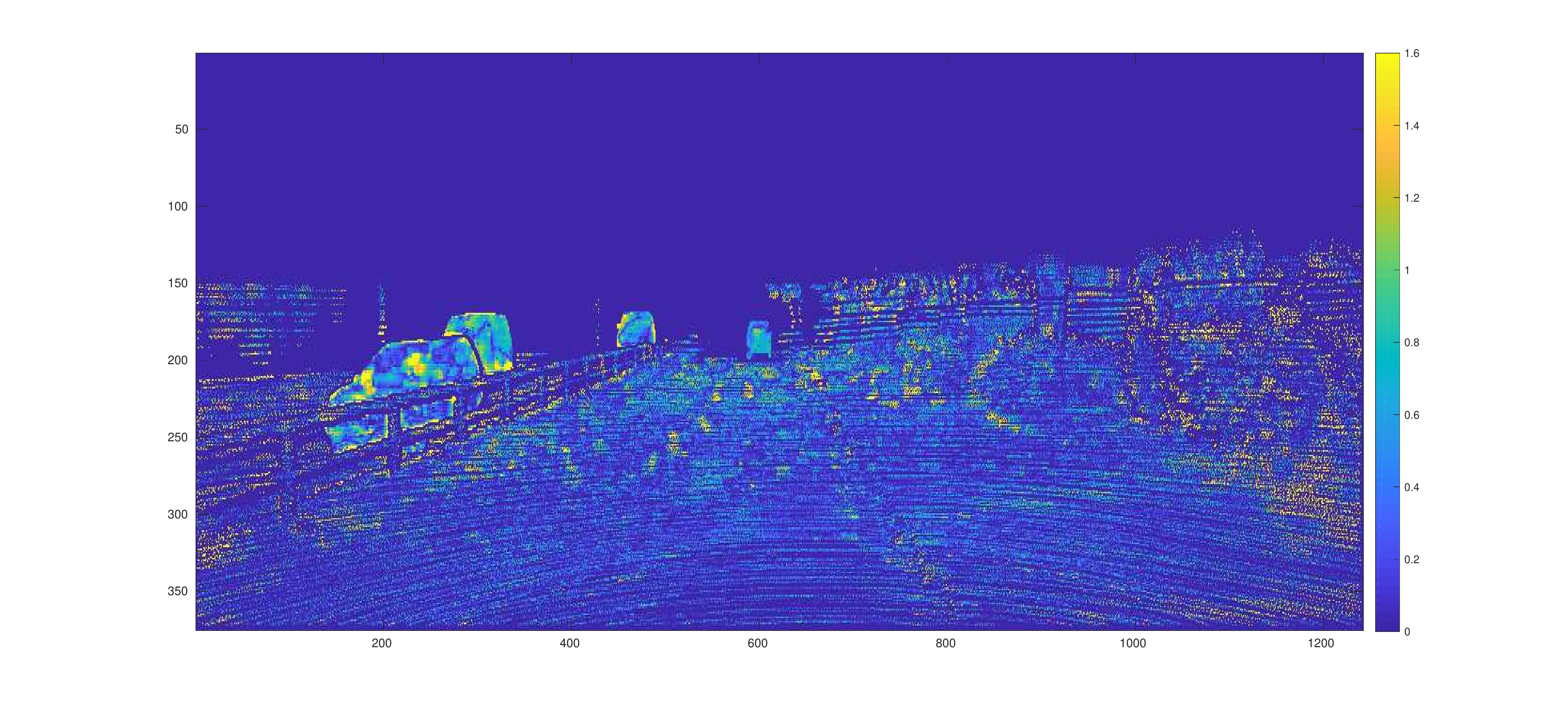}}  \\		
	
	\caption{The unsupervised adversarial network was trained  to fuse the initial disparity maps (c), (e) into a refined disparity map (g) for the same scene (b) from the Kitti2015 dataset~\cite{Kitti2015}. (a) is the corresponding ground truth. (d), (f), (h) are the errors of (c), (e), (g). The colorbars (from blue to white) corresponds to 0 - 1.6 pixels and the lighter pixel have bigger error in (d), (f), (h).}
	
	\label{fig:example_demo_UDFNet}
\end{figure*}

For qualitative results, see Figure~\ref{fig:example_demo_UDFNet}. The proposed method compensates for the weaknesses of the inputs and refines the initial disparity maps effectively. Compared with SGM~\cite{SGM}, the fused results are much more dense and accurate. Compared with PSMNet, the proposed method preserves the details better (eg: mountain). But the proposed method fully fails in the sky. The reason is that the pixels (disparity = 0) are treated as invalid (confidence = 0) in SGM. However, the disparity values of the pixels in the sky area from PSMNet are all larger than 0 (confidence \textgreater 0). So, the PSMNet misleads the network to adopt their disparity value as the initialization. Thus, the wrong confidence measurement can bring big error to the refined disparity map. The problem can be solved by adding more cues, such as semantic meaning, to make the confidence measurement more accurate.

The efficiency and effectiveness of the proposed network's structure on the Kitti2015 dataset have been explored. The same settings from Table~\ref{Network_setting_UDFNet} except the channel ($l_g$, $l_d$) of the features in each layer were used. The results are shown in Table~\ref{Network_structure_efficiency_UDFNet}. When $l_g=l_d=10$, our network achieves the best accuracy (0.83 pixels) compared with the rest. Additionally, the four groups of experiments with different $l_g, l_d$ achieve similar performance, which shows the robustness of the proposed method.  

\begin{table}[h!]
	\centering
	\caption{Average error (pixel) on Kitti2015 } \label{Network_structure_efficiency_UDFNet}	
	\begin{tabular}{|p{3cm}|*{2}{p{2.6 cm}|}}  % repeats {c|} 6 times
		\hline
		Structure  &  Error (pixel) &  Time (s/frame)   \\ \hline 
		$l_g=l_d=5$  & 1.00 & \bf 0.009   \\ \hline
		$l_g=l_d=10$  & \bf 0.83 & 0.011   \\ \hline
		$l_g=l_d=15$  & 0.92 & 0.014   \\ \hline
		$l_g=l_d=20$  & 0.87 & 0.017   \\ \hline						
	\end{tabular}	
\end{table}

\subsubsection{Stereo-Lidar Fusion}
Our network can generalize to any sub-fusion task based on left-right consistency. A stereo-lidar fusion demo on Kitti2015 has also been done. The valid points in the initial ground truth were removed by half randomly to get the Lidar input to our network. The confidence of each valid Lidar data point was set as extremely large (100\%).  PSMNet was used as the stereo input again and set the confidence of each pixel as 50\%. One hundred labeled images in the initial Kitti2015 training dataset were used to train (from `000100\_10.png' to `000199\_10.png' ) and the other 100 to test (from `000000\_10.png' to `000099\_10.png'). The error of PSMNet~\cite{PSMNet} is 1.22 pixels and our error is 0.86 pixels after fusion.

\section{Conclusion}
We proposed an unsupervised method to fuse the disparity estimates of multiple state-of-art disparity/depth algorithms. The experiments have shown the effectiveness of the energy function design based on multiple cues and the efficiency of the network structure. The proposed network can be generalized to other fusion tasks based on left-right image consistency (In this paper, we only did stereo-stereo and stereo-lidar fusion). The work in this paper reduces the cost of acquiring labelled data necessary for use in a supervised method. Given the algorithm's low computation cost, the combination of the proposed method and existing depth-acquisition algorithms is a good solution to obtaining high accuracy depth maps. Future work will investigate improved methods for setting the confidence values based on the initial disparity values and type of sensor. 

\bibliographystyle{unsrt}  
\bibliography{references}  %%% Remove comment to use the external .bib file (using bibtex).
%\begin{thebibliography}{1}

\end{document}